\pdfoutput=1
\documentclass[11pt]{article}

\usepackage{ACL2023}
\usepackage{enumitem}
\usepackage{times}
\usepackage{latexsym}
\usepackage{pgfplots}
\usepgfplotslibrary{groupplots}
\usetikzlibrary{matrix}
\usepackage[T1]{fontenc}
\usepackage{changepage}
\usepackage[utf8]{inputenc}

\usepackage{microtype}
\usepackage{graphicx}
\usepackage{inconsolata}
\usepackage{amsmath}
\usepackage{amsfonts}
\usepackage{algorithm}
\usepackage{algorithmic}
\usepackage{multirow}

\title{Eliminating Reasoning via Inferring with Planning: A New Framework to Guide LLMs' Non-linear Thinking}

\author{
  Yongqi Tong\textsuperscript{1}, 
  Yifan Wang\textsuperscript{2}, 
  Dawei Li\textsuperscript{1}, 
  Sizhe Wang\textsuperscript{3}, 
  Zi Lin\textsuperscript{1}, 
  Simeng Han\textsuperscript{4}, 
  Jingbo Shang\textsuperscript{1}\textsuperscript{\textdagger} \\
  \textsuperscript{1}University of California, San Diego, \texttt{\{yotong, dal034, lzi, jshang\}@ucsd.edu} \\
  \textsuperscript{2}University of Pennsylvania, \texttt{yyifan@seas.upenn.edu} \\
  \textsuperscript{3}University of Southern California, \texttt{sizhewan@usc.edu} \\
  \textsuperscript{4}Yale University, \texttt{simeng.han@yale.edu}
}

\begin{document}
\maketitle
\renewcommand\thefootnote{\textdagger}
\footnotetext{Corresponding author.}
\begin{abstract}
% As the overall capabilities of large language models (LLMs) vastly expand, recent work are now pivoting towards equipping LLMs with human-like reasoning abilities. To further align their reasoning processes with those of humans, it becomes imperative to find guidance that better direct LLMs with high-level human cognition and logic.
% Many recent studies explore equipping large language models (LLMs) with high-level reasoning abilities by emulating human-like cognition and logic.
Chain-of-Thought(CoT) prompting and its variants explore equipping large language models (LLMs) with high-level reasoning abilities by emulating human-like linear cognition and logic. 
However, the human mind is complicated and mixed with both linear and nonlinear thinking. 
In this work, we propose \textbf{I}nferential \textbf{E}xclusion \textbf{P}rompting (IEP), a novel prompting that combines the principles of elimination and inference in order to guide LLMs to think non-linearly. 
IEP guides LLMs to plan and then utilize Natural Language Inference (NLI) to deduce each possible solution's entailment relation with context, commonsense, or facts, therefore yielding a broader perspective by thinking back for inferring. 
This forward planning and backward eliminating process allows IEP to better simulate the complex human thinking processes compared to other CoT-based methods, which only reflect linear cognitive processes.
We conducted a series of empirical studies and have corroborated that IEP consistently outperforms CoT across various tasks. 
Additionally, we observe that integrating IEP and CoT further improves the LLMs' performance on certain tasks, highlighting the necessity of equipping LLMs with mixed logic processes.
Moreover, to better evaluate comprehensive features inherent in human logic, we introduce \textbf{M}ental-\textbf{A}bility \textbf{R}easoning \textbf{B}enchmark (MARB). The benchmark comprises six novel subtasks with a total of 9,115 questions, among which 1,685 are developed with hand-crafted rationale references. 
We believe both \textsc{IEP} and \textsc{MARB} can serve as a promising direction for unveiling LLMs' logic and verbal reasoning abilities and drive further advancements. \textsc{MARB} will be available at ~\texttt{anonymity link} soon.

\end{abstract}

\section{Introduction}

\begin{figure}[htbp]

    \centering
    \scalebox{0.42}{\includegraphics{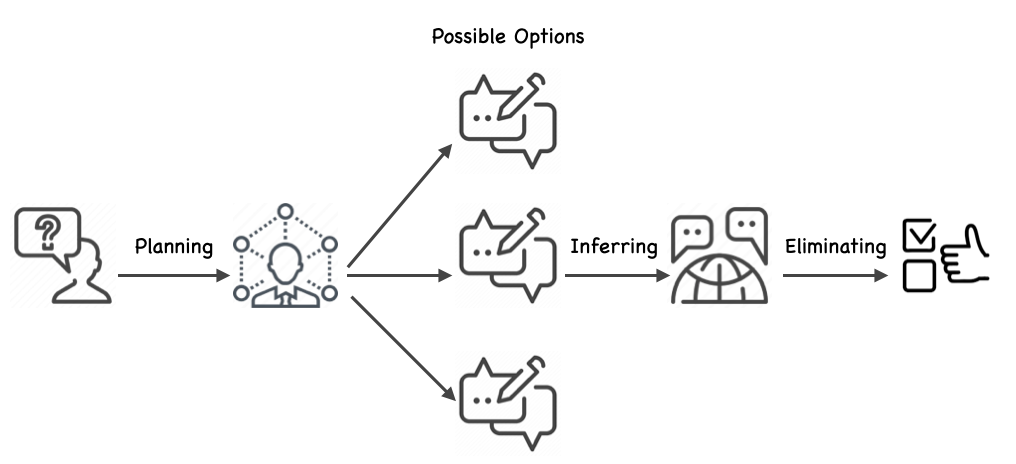}}
    \scalebox{0.6}{\includegraphics{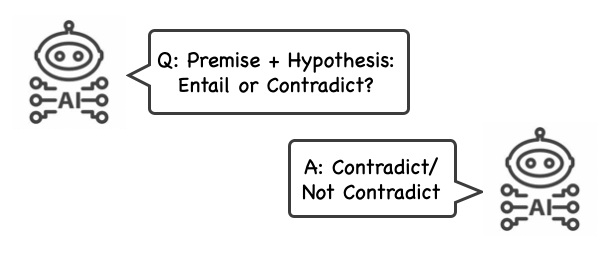}}
    \caption{Our methods, Inferential Exclusion Prompting, using elimination skills to arrive at the final answer. 
    Our approach to guiding LLMs involves encouraging them to consider and plan multiple possible options, while also aligning their natural language inference skills to double-check whether each option is aligned with the context and common sense. }
    \label{fig: IEP steps}
    \vspace{-0.5cm}
\end{figure}

Recently, the rapid development of large language models (LLMs)~\citep{brown2020language,zhang2022opt,anil2023palm,touvron2023llama} and their striking skill and knowledge have sparked significant interest in equipping them with human-like reasoning abilities~\citep{patel2021nlp,kojima2022large,wang2023boosting}.
To do so, some works~\citet{wei2022chain,wang2022self} design inferential pipelines for LLMs by emulating human's Chain-of-Thought (CoT) thinking and achieve promising results. 
% \simeng{Should we add comparison to Tree-of-Thought?}
However, the human cognitive system is sophisticated and multifaceted~\citep{stenning2012human,knauff2010complex,mercier2011humans}, which does not only rely on such linear thinking to solve problems in reality.
For example, when confronted with intricate issues, humans commonly employ intuitive divergent or reverse thinking to explore various solutions, starting with the desired or intuitive outcome and tracing backward to pinpoint prerequisites or steps. 
This sophisticated nature of human thinking highlights the importance of exploring more refined reasoning paradigms to improve LLMs' inference capability further.

% Elimination method represents a cognitive approach where one begins by understanding, planning and then determining the correct answer by systematically ruling out improbable or irrelevant options based on the given context or prior knowledge~\cite{de1996levels}. After enumerating all possible solutions, the process narrows down to discern the relatively optimal solution, making the decision-making more systematic and efficient. Intriguingly, this cognitive method can be likened to a form of Natural Language Inference (NLI) task. When leveraging Large Language Models (LLMs), we can prompt them to infer each option explicitly and determine its alignment or contradiction with the provided context or commonsense. 

In this study, we introduce Inferential Exclusion Prompting (IEP), inspired by the elimination method in logic. This thinking methodology is formally known as \textit{disjunctive syllogism} in cognitive science~\citep{hurley2014concise}. 
The elimination method represents an approach where one begins by understanding, planning, and then determining the correct answer by systematically ruling out improbable or irrelevant options based on the given context or prior knowledge~\citep{de1996levels}. 
As Figure~\ref{fig: IEP steps} shows, given a specific problem, IEP initially instructs LLMs to plan possible answers. 
Subsequently, IEP inventively models the elimination of each candidate choice as an NLI process. 
This enables LLMs to leverage their proficiency in NLI for elimination, thereby fully promoting their potential in the reasoning process. 
By making determinations for each candidate, IEP can effectively exclude incorrect options and identify the correct answer.

In Section~\ref{sec:exp}, through extensive evaluation across several reasoning tasks, we affirm that IEP consistently outperforms original CoT-based methods. This emphasizes the significance of non-linear thinking in IEP. 
Furthermore, we observe additional performance enhancements when combining IEP with CoT. 
This confirms that IEP is highly compatible with other reasoning frameworks in building more sophisticated and human-like inference systems.

% Moreover, LLMs has skyrocketed to an marvelous accuracy of 99.9\%~\cite{wang2023boosting} on some existing benchmarks. As the amount of experience LLMs have learned continues to increase, a common issue in most current NLP tasks is the insufficient challenge presented by existing benchmarks. Therefore, we believe building more in-depth and sophisticated human reasoning benchmarks has become a stepping stone to the future. Hence, we introduce \textsc{MARB}, a comprehensive and challenging benchmark to represent the innovation and generality inherent in human logic. As a comparative dataset to evaluate the similarities between human logic and LLMs' reasoning process, \textsc{MARB} has been enriched with a collection of human rationale references, which is demonstrated to be efficacious for the following work in this study. This benchmark draws from a diverse set of challenges, inclusive of parajumbles, riddles, puzzles, brain teasers and critical reasoning queries. We have conducted a series of empirical studies to ensure its latent significance and potential as a future direction for enhancing LLMs' logical reasoning skills. Some representative examples can be found in Fig~\ref{fig:benchmarks} and more are displayed in the Appendix~\ref{app:examples}.

In order to further expose humans' non-linear thinking pattern in complex verbal reasoning tasks, we collect a comprehensive and challenging benchmark, \textbf{M}ental-\textbf{A}bility \textbf{R}easoning \textbf{B}enchmark (MARB), to evaluate our IEP thoroughly with various kinds of reasoning pattern. 
\textsc{MARB} draws from diverse games and challenges, including parajumbles, riddles, puzzles, brain teasers, and critical reasoning queries.
The task of parajumble, where LLMs are required to rearrange jumbled sentences into a coherent paragraph, is an unprecedented paradigm for both Natural Language Understanding(NLU) and parallel reasoning. In this novel assignment, LLMs grasp global information and demand a high-level understanding of the logical relationship among individual sentences. 
Furthermore, \textsc{MARB} has been enriched with a collection of human rationale references. Section~\ref{sec:human-reference} demonstrated this part enables future work to analyze the similarities between human logic and LLMs' reasoning process in a fine-grained manner.
Figure~\ref{fig:benchmarks} presents several representative examples in \textsc{MARB}.
We will release all the datasets and codes to provide a direction for future efforts in equipping LLMs' reasoning abilities to learn from diversified human logic.

In summary, our contributions to this work are bi-fold: 

\begin{itemize}[itemsep=1.5pt]
    \item We propose a human-like logical prompting method, IEP, that simulates human elimination-based reasoning. It splits the reasoning into several steps and novelly models the elimination problem as an NLI task. Through extensive experiments, we validate the effectiveness of IEP to outperform other CoT-based methods consistently. 
    \item We publish \textsc{MARB}, a comprehensive dataset encapsulating a diverse set of reasoning games and challenges. The inclusion of hand-crafted human rationale provides deeper insights into human decision-making processes. Some examples from \textsc{MARB} are shown in Appendix~\ref{app:examples}. 
    % \item We also conduct further analysis to demonstrate the strong compatibility of our IEP with various reasoning frameworks \lzi{add some details}.
\end{itemize}

\section{Motivation}
\label{sec:cot-problems}
In this section, we primarily analyze the main drawbacks of CoT-based prompting methods and present the necessity of our framework. 

\textbf{Error Propagation}~
CoT aims to describe the chain-like structure and linear thinking in human cognition, where one step of thought is followed by considering the next step. This is very intuitive, drawing inspiration from a classic method of human thinking. However, the problem with this approach is that any minor error in an intermediate step can propagate through the chain and lead to the failure of subsequent tokens generated by LLMs. Formula~\ref{eq:CoT_formula} illustrates how minor inaccuracies can snowball into more significant issues as the chain progresses. Figure~\ref{table:mistake} shows one CoT's mistake guided by the initial step.

\begin{equation}
\begin{aligned}[b]
&L(x_1,  \dots, x_n | \boldsymbol{\phi}, e_1, \dots, e_{n-1}) = \\
&P(x_1 | \boldsymbol{\phi}) \times \prod_{i=p+1}^{n} P(x_i | x_{i-1}, \dots, x_{i-p}, \boldsymbol{\phi}, e_{i-1})
\end{aligned}
\label{eq:CoT_formula}
\end{equation}
where \( x_1, \ldots, x_n \) denote the steps or links in the chain of thought, \( \boldsymbol{\phi} \) represents the LLM's parameters, \( e_1, \ldots, e_{n-1} \) encapsulate the error factors at each step.

Although some previous works propose an uncertainty score to rethink the uncertain intermediate steps~\citep{jiang2023active} in order to fix this problem, we still lack an effective solution that can handle these inherent problems.

\begin{table}[h]
\centering
\begin{tabular}{p{0.9\columnwidth}}
\hline
\textbf{Reorder these sentences to make it coherent and logically correct.} \\
\hspace{1em} \textbf{Options}: A: As this is a sunny afternoon, \\
\hspace{1em} he decided to take a walk in the park. \\
\hspace{1em} B: He sat on the couch, completely \\
\hspace{1em} engrossed in his favorite TV show. \\
\hspace{1em} C: In the park, he unexpectedly ran into an  \\
\hspace{1em} old friend and they chatted happily. \\
\hspace{1em} \textbf{Correct Answer}: B → A → C \\
\hline
\textbf{COT Reasoning:} \\
\hspace{1em} \textbf{Step1} The best possible start is A because\\
\hspace{1em} it introduces the background information.\\
\hspace{1em} \textbf{Step2} Then most likely next step is C\\
\hspace{1em} because its location is a park. \\
\hspace{1em} \textbf{Step3} Then B is the last sentence. \\

\hspace{1em} So the final answer is: A → C → B \\
\hline
\textbf{IEP Reasoning(Our Method):} \\
\hspace{1em} \textbf{Step1} All possible options are: A → B →\\
\hspace{1em} C, A → C → B, ... \\
\hspace{1em} \textbf{Step2}  A → C → B is incorrect because B\\
\hspace{1em} is inconsistent with the other ones without \\ 
\hspace{1em} any turning point. \\
\hspace{1em} ... \\
\hspace{1em} So the final answer is: B → A → C \\
\hline
\end{tabular}
\caption{An example where CoT guides PaLM2 to make a mistake.}
\label{table:mistake}
\end{table}

% \begin{figure}[h]
%     \centering
%     \scalebox{0.55}{\includegraphics{pic/comparisons.png}}
%     \caption{An example where CoT guides PaLM2 to make a mistake.}
%     \label{fig:mistake}
% \end{figure}

\textbf{Singular Thinking}~
CoT naively underscores the sequential and step-by-step progression of thoughts. While such an approach ensures systematic analysis, it fosters singular thinking, limiting exploration to a non-linear perspective or general situations. Consequently, this narrowed focus might neglect the consideration of alternative insights or multifaceted strategies to address complex issues. 

\citet{zhou2022least} requests LLMs to generate multiple chains to consider all possible solutions, but a simple majority vote determines the final result. \citet{yao2023tree, yao2023beyond} propose to use some complex structures, such as trees and graphs, to address the inherent limitations of chains. Unfortunately, most of those solutions incur a significant increase in computational resources. 

\textbf{Internal Evaluation Difficulties}~
CoT prompting poses challenges in the evaluation of its intermediate steps. The current approach is to evaluate the final answer directly. However, irrelevant or even counterfactual steps existing in the intermediate stages remain undetectable, which raises concerns about the reliability and robustness of the reasoning. Although \citet{prasad2023receval} proposed using a pre-trained NLI model to evaluate the CoT under certain hand-crafted factors, enhancing the CoT reasoning capabilities of LLMs remains a practical challenge.

% As discussed above, those subsequent CoT-based promptings are excessively complex, deviating from the natural human cognitive process. We still need more promptings or guidance frameworks inherited in human logic to ensure LLMs' structured, deductive, and trustful reasoning process without excessive resource consumption.

In summary, current CoT-based promptings still have limitations in various aspects. To address these challenges and fully leverage the potential of LLMs to perform reasoning tasks, we propose the Inferential Elimination Prompting (IEP) framework.

\section{Our Methods: Inferential Elimination Prompting (IEP) Framework}
\label{method}
In this section, we introduce IEP, a novel prompting to guide LLMs to explicitly infer each provided or pre-planned option based on the given context and commonsense facts. 
As opposed to CoT, IEP is inherently capable of forward planning several possible answers to make a comprehensive deduction and then backward checking those options. 
IEP aims to use Pretrained Language Models' capabilities to reason by exclusion. The process can be divided into three main steps.
In step 1, IEP first guides LLMs to learn the whole context and generate possible answers. Step 2 is an inferring template to explicitly infer each option's relation with the context or facts. This step will eliminate those contradicted or disinherited options and hence narrow the scope. In step 3, rethink the possible answers to choose the best one or all that apply.

% The guidance framework naively avoids those problems mentioned in Section~\ref{sec:cot-problems}. One possible concern for error propagation in the process between planning and inferring can be addressed because, in the first step, LLMs propose many potential scenarios, and in the second step, the inference naturally eliminates those incorrect ones within LLMs' possible capabilities. The detailed empirical studies will later be discussed in Section~\ref{sec:exp}. 

\begin{algorithm}
    \caption{IEP}
    \begin{algorithmic}
        \STATE \( \textit{candidate} = \{\} \)
        \STATE \textbf{Step1: }Planning
        \STATE \textbf{Prompt:} Understand the problem and propose k possible answers.
        \STATE \textit{candidate} \( \leftarrow \) k possible answers
        
        \FOR{option in \textit{candidate}}
            \STATE \textbf{Step2: }Inferring
            \STATE \textbf{Prompt:} What is the premise if it is true?
            \STATE \textbf{Optional Step: }Self-ask or retrieve relevant information as the options' premise. 
            \STATE \textbf{Prompt:} Think the option as hypothesis. Does it entail with those premises?  
            \STATE \textbf{Step3: }Eliminating
            \IF{Entail}
                \STATE Continue
            \ELSIF{Contradict}
                \STATE Remove from \textit{candidate}

            \ENDIF
        \ENDFOR
        \STATE \textbf{Step4: }Answering
        \STATE \textbf{Prompt: }So the final answer is: 
    \end{algorithmic}
\end{algorithm}

\subsection{Planning}
Before proceeding with elimination, it is essential to establish the boundaries for decision-making under the provided context, allowing for more precise elimination in the subsequent steps. 
Although most existing reasoning benchmarks will provide the possible options in real-world scenarios, especially during interactions between LLMs and users, the tendency is for the LLMs to generate their own answers rather than relying on options provided by the user. This step is foundational and sets the stage for the success of the IEP's overall strategy. 

Given a context denoted as \( C \), a pre-trained language model represented as \( PLM \) is used to generate a set of potential answers or options, represented as \( A \).

\begin{equation}
A = PLM_{\text{gen}}(C)
\end{equation}

Here, \( A \) consists of the potential answers: \( A = \{a_1, a_2, ... , a_n\} \).

\subsection{Inferring}
Inferring determines the validity of each option relative to the provided context or commonsense knowledge. This step involves simple pattern recognition and a deeper understanding and connection of the option with prevailing facts and premises. In this phase, the LLM is prompted to consider the candidate's relation with context. The option will be voted out if it is not aligned. 

The prompt "What is the premise if it is true?" is an example to guide the LLM to identify and evaluate these premises for each option. If every identified premise for an option aligns with the commonsense, given facts and context, then the option can be inferred as true or valid. Conversely, even if a single premise is found to be untrue or inconsistent, the associated option becomes logically untenable.

For each premise \( p_{ij} \), using the PLM, we assess its consistency with the context \( C \) and obtain a binary score \( s_{ij} \).

\begin{equation}
s_{ij} = PLM_{\text{infer}}(p_{ij}, C)
\end{equation}

To compute the overall consistency score \( s_i \) for the answer \( a_i \), we calculate the product of all the binary scores for its associated premises:

\begin{equation}
s_i = \prod_{j} s_{ij}
\end{equation}

Given the binary nature of \( s_{ij} \), the product \( s_i \) will be 1 if and only if all premises for \( a_i \) are consistent with \( C \); otherwise, it will be 0

\subsection{Eliminating}

The optimal answer is the one with the highest score, \( a^* \):

\begin{equation}
a^* = \arg\max_{a_i \in A} s_i
\end{equation}
By integrating the steps, the primary aim of IEP is to identify the answer(s) that resonate the most with the given context:

\begin{equation}
a^* = \arg\max_{a_i \in PLM_{\text{gen}}(C)} PLM_{\text{inf}}(C, a_i)
\end{equation}

\subsection{Integrating other prompting skills}
Pure logical evaluation, while powerful, might sometimes be restrictive given the vast complexities and nuances of real-world scenarios. 
To address this and to aid the LLM in thinking more comprehensively and accurately, IEP can integrate CoT, self-ask prompting~\cite{huang2022large}, or active retrieval-augmented prompting~\cite{diao2023active}. 

When making decisions and eliminating possible options, people will also refer to various sources of information. Combining CoT and IEP might help LLMs make better decisions on some tasks in Section~ ef {sec:exp}. By integrating other prompting techniques, IEP can guide LLMs to reason as humans do. 

\section{Construction of \textsc{MARB}}
As highlighted in Section~\ref{method} and Section~\ref{related-work}, IEP is oriented towards complex reasoning. Recent work on the constructions of human-like complex reasoning benchmark focuses on arithmetic reasoning~\citep{cobbe2021training,patel2021nlp,ling2017program}, which primarily focused on linear thinking and hence CoT achieves state-of-the-art performance on that~\citep{wei2022chain}. 
Unfortunately, we are still lacking a complex verbal reasoning benchmark to reflect the human's logic and reasoning abilities. In this section, we introduce \textsc{MARB}, a diversified and challenging benchmark designed to reflect the innovative facets of human logic and linguistic comprehension. Our dataset is built on these domains. 

\subsection{Parajumbles}
In the realm of human logic challenges, parajumbles, also known as sentence reordering, stands out not merely as a linguistic task but as a potent evaluator for text understanding and human reasoning. Parajumbles involve rearranging jumbled sentences to form a coherent and logically sequenced paragraph. The inspiration for this task is drawn from two well-known tests in their local community - Common Admission Test(CAT)\footnote{\url{https://cdn.digialm.com/EForms/configuredHtml/756/84433/Registration.html}} and Pearson Test of English for Academic(PTE)\footnote{\url{https://www.pearsonpte.com/}}. These assessments are tailored for entry examinations at the undergraduate or graduate academic levels, especially for science, engineering, and business major students. This task is by no means trivial, even for humans. For instance, according to the descriptions on PTE's official website, in the 2019 PTE examination, out of a maximum score of 90, the median was only 62.57 points. Besides CAT and PTE, we also collect and shuffle those paragraphs from~\cite{misra2022news,harinatha2021evaluating}, two open-sourced news datasets collected from various corpora, such as HuffPost, Business Insider, and CNN. 

Success in these tasks requires a deep understanding of the causal, concurrent, and structured relationships and the implicit logic inherent in human texts. While it is similar to the Next-Sentence-Prediction(NSP) task, parajumble demands a deeper understanding of the global context. NSP probes for localized coherence between two sentences, but parajumble is more challenging as it requires delving into global coherence across a set of sentences. This task can be an unprecedented direction for LLMs' finetuning and reasoning in the next phase. 

\subsection{Puzzles \& Brain Teasers}
Puzzles and brain teasers are more than intricate games or intellectual diversions. By nature, they are designed to challenge our cognitive faculties, forcing us to tap into both learned knowledge and innate logic in real-world problems. Unlike riddles, which play on linguistic ambiguities or parajumbles, focusing on reconstructing logically coherent narratives, Puzzles and brain teasers hinge on methodical, step-by-step deduction and inference of structured problems. We collect puzzles and brain teasers from sawaal\footnote{\url{https://www.sawaal.com/}}, a well-known public website. All contents, including human rationale, have long been checked and reviewed by the public. 

The primary goal for this subset reflects human's inherent mental abilities to break down a problem,  identify or follow its underlying patterns, utilize prior knowledge to solve a problem step by step, and then reconstruct a solution. This iterative process of deconstruction and reconstruction is a testament to our ability to adapt and apply structured logic in varying contexts. 

\subsection{Riddles}
Much like parajumbles, riddles also manifest as a complex reflection of the human reasoning process. However, what sets riddles apart is their emphasis on linguistic creativity and the play of words. These challenges compel individuals to navigate through deceptive intricacies, metaphors, and allusions. They are designed to provoke thought, elicit surprise, and often lead the solver down unexpected paths of contemplation. 
Learning from this subset will require LLMs to grasp hidden meanings and toggle between different interpretations of words or phrases to truly understand the ambiguity of language and hence promote tricky and unexpected cognitive realms. The primary intent of this task is to compel LLMs to think beyond the immediate context, pushing them to derive insights and formulate responses that are not strictly confined to the provided information, which inherits humans' intrinsic ability in reasoning. We collect those well-designed riddles from an open-sourced website famous for stimulating cognitive explosions, ahaPuzzles\footnote{\url{https://www.ahapuzzles.com/}}. 

Although some previous works have touched upon the subject, they have superficially lacked the depth to encapsulate intrinsic and distinct human reasoning. For example, while ~\cite{lin2021riddlesense} introduced innovative approaches to utilize riddles to represent human logic, its single topic raises concerns for being excessively tricky given the riddles' emphasis on ambiguity, metaphor, and wordplay. Moreover, we have also designed a human rationale to enhance the performance of LLMs.

\subsection{Critical Reasoning}
The fourth part of \textsc{MARB} that reflects human reasoning logic is critical reasoning, which stands as a pivotal component in assessing advanced human cognition. Inspired by GRE's\footnote{\url{https://www.ets.org/gre.html}} and GMAT's\footnote{\url{https://www.mba.com/exams/gmat-exam/}} creative reasoning questions, this segment is intended to reflect LLMs' understanding and reasoning abilities about paradox, assumptions and conclusions rather than an essential Natural Language Understanding(NLU) task. These facets underscore human logic's essence, characterized by its sequentially interlinked nature and its inherent capacity for being computationally abstract and deductible.  

Although the question format of MARB closely aligns with that of ReClor as discussed in ~\cite{yu2020reclor}, there are salient differences between the two. Primarily, a significant portion (84.54\%) of the MARB dataset encompasses rationale references provided by educators who have years of experience teaching these graduate admission examinations, which is absent in ReClor. Additionally, we've meticulously curated our dataset by eliminating entries with identical option content in ReClor, further refining its uniqueness.
% \subsection{Analysis on \textsc{MARB}}

\section{Experiments}
\label{sec:exp}
Using the proposed \textsc{MARB} and some existing benchmarks, we next conduct a series of experiments and analysis on different LLMs. 

Here are existing datsets that we choose. 

\begin{itemize}[itemsep=0pt]
    \item CommonsenseQA~\cite{talmor2018commonsenseqa} benchmark of multiple-choice questions that require different types of commonsense knowledge to obtain the correct answers. 
    \item OpenbookQA~\cite{mihaylov2018can} dataset collected from some elementary-level science facts and also designed with multiple choices. 
    \item StrategyQA~\cite{geva-etal-2021-aristotle} benchmark dataset with questions requiring multi-step reasoning. 
    \item LogiQA~\cite{liu2020logiqa} covers multiple types of deductive reasoning to investigate the logical reasoning abilities of LLMs.
\end{itemize}

We compare our method with Zero-shot Standard Prompting~\cite{kojima2022large} and CoT Prompting~\cite{wei2022chain}. For LLMs, we consider two of the strongest baselines used for reasoning, GPT4~\cite{openai2023gpt4} and PaLM2-540B~\cite{anil2023palm}. 

\subsection{IEP's Results and Analysis}

\begin{table*}[h]
    \centering
    \begin{tabular}{c|cccc}
    \hline \textbf{Method}     & \textbf{OpenbookQA} & \textbf{StrategyQA} & \textbf{CommonsenseQA} & \textbf{LogiQA}
    \\\hline
    Standard Prompting  & 80.92   & 91.23    & 74.17 & 42.21 \\
    CoT & 82.66   & 91.86    & 76.32 & \underline{41.05} \\\hline
    IEP & \textbf{88.98} & 92.24 & 77.52 & \textbf{45.75}\\
    IEP$\cup$CoT & 85.33 & \textbf{93.25} & \textbf{78.41} & 42.74\\\hline
    \end{tabular}
    \caption{PaLM2-540B's accuracy on existing benchmarks via different prompting skills. CoT$\cup$IEP represents the integration of IEP and CoT, systematically eliminating irrelevant considerations step by step. Underlined sections underscore where CoT has a detrimental effect compared with Standard Prompting reasoning.}
    \label{tab:existing benchmarks performance}
\end{table*}

\begin{table*}[h]
    \centering
    \begin{tabular}{c|c|cccc}
    \hline \textbf{Model} & \textbf{Method}     & \textbf{Puzzles} & \textbf{Riddles} & \textbf{Parajumble} & \textbf{CR} \\\hline
    \multirow{4}{*}{PaLM2 540B} & Standard Prompting       & 49.45   & 61.90  & 25.54 &  58.39 \\
                          & Zero-Shot CoT   & 53.24   &  63.03 & \underline{20.08} &  \underline{51.98} \\\cline{2-6}
                          % & 8-Shot CoT      &   -  & - & - &  - \\
                          & IEP             & 57.28 & 62.78 & \textbf{27.84} & \textbf{61.27} \\
                          & IEP$\cup$CoT    & \textbf{60.02} & \textbf{63.43} & 23.62 & 56.75\\\hline\hline

    \multirow{4}{*}{GPT4} & Standard Prompting       &  64.37 & 67.70 & 52.17 &  65.32 \\
                          & Zero-Shot CoT   & 81.22   &  \textbf{81.92} & \underline{45.96} & \underline{63.01}  \\\cline{2-6}
                          & IEP             & 79.32 & 78.30 & \textbf{67.85} & \textbf{69.24} \\
                          & IEP$\cup$CoT    & \textbf{82.03} & 79.28 & 51.94 & 65.73 \\\hline
    \end{tabular}
    \caption{PaLM2-540B and GPT4's accuracy on \textsc{MARB}. Puzzles include puzzles and brain teasers. CR stands for critical reasoning subset. }
    \label{tab:marb performance}
\end{table*}

Table~\ref{tab:existing benchmarks performance} presents PaLM2-540B's evaluation results on existing benchmarks. IEP and the combination of both IEP and CoT significantly outperform the other two promptings. Especially on OpenbookQA, our method outperformed CoT by 6.32\% without any additional computational resource consumption. One of the dominant reasons is that OpenbookQA explicitly specifies the relevant commonsense and facts, which directly provides a crucial premise for the NLU task. This suggests that IEP is adept at handling tasks that involve such facts. CoT's performance is notably mixed. In some instances (such as on LogiQA), it underperforms compared to the zero-shot baseline. This underscores a potential limitation or misalignment of CoT's approach with certain tasks. 

The combination of IEP and CoT generally shows improvements, but it does not always outperform IEP alone. 
This suggests that while the combination has its benefits, there are scenarios where adding CoT's approach might not be beneficial.
One of the potential reasons is the inventive alignment from commonsense or complex reasoning tasks to NLI tasks. 
A lot of previous work has optimized LLMs on their capabilities of NLI tasks. 
However, few people have applied the CoT task to NLI, which might confuse LLMs due to the different nature of the two tasks. 
% \simeng{Maybe clarify more on what the benefits are, the causes of them, and how CoT might not be beneficial.}

Table~\ref{tab:marb performance} shows the performance of the two LLMs on \textsc{MARB}. As we can see, CoT struggles with the parajumble task. The reason might be that parajumble largely tests concurrent reasoning, where one hypothesizes a sequence and then thinks in reverse to verify its correctness. CoT's step-by-step thinking approach can easily introduce errors at the very beginning of the logic. As we can see in Table~\ref{table:mistake}, IEP can make LLMs think globally and in reverse to verify whether each candidate's logic is aligned with the global context.

However, we notice that CoT may exhibit superior performance over IEP when handling riddle tasks. It is reasonable as riddles often emphasize moments of eureka or sudden insight in human cognition, which may not align well with the meticulous and structured inference characteristic of IEP. 

\subsection{Utility of Human Reference in \textsc{MARB}}
\label{sec:human-reference}

In this section, we conduct k-shot In-Context Learning(ICL) experiments evaluating the utility of human rationale in \textsc{MARB}. 

\begin{table}[h]
    \centering
    \scalebox{0.85}{
    \begin{tabular}{c|cccc} \hline
      \textbf{Method}     & \textbf{Puzzles} & \textbf{Riddles} & \textbf{Parajumble} & \textbf{CR} \\\hline
         0 shot       &  81.22   &  81.92 & 45.96 & 65.32 \\
         1 shot & 82.92 & 80.53 & 46.27 & 65.97\\
         8 shot & 84.90  & 85.63 & 51.42 &  68.73 \\\hline
    \end{tabular}
    }
    \caption{GPT4's k-shot ICL performance on \textsc{MARB}. }
    \label{tab:my_label}
\end{table}

As the number of shots (or training examples) increases, the performance across most tasks seems to improve. Specifically, for the Puzzles and Riddles tasks, there's a noticeable increase in performance from the 0-shot to the 8-shot learning. The Parajumble task, though starting with a lower performance score, also shows a similar positive trend. 

The evaluation showcases the utility of human reference in \textsc{MARB}. It is evident that increasing the number of shots or examples benefits the model's accuracy, especially in tasks like Puzzles, Riddles, and Parajumble. This analysis suggests that for tasks demanding a deeper understanding or complex reasoning, a higher number of shots might provide better guidance to the model, leading to improved outcome.

\section{Related Work}
\label{related-work}
% \lzi{I suggest use a table to summarize different kind of reasoning tasks, and also for the final MARB benchmark}

As mentioned by ~\citet{xu2023large,bang2023multitask,yu2023nature}, existing common reasoning tasks and related benchmarks can be categorized into several distinct groups. Table~\ref{tab:reason} shows the main categories. 

\begin{table*}[h]
    \centering
    \scalebox{0.8}{
    \begin{tabular}{p{5cm}|p{6cm}|p{6cm}}\hline
     \textbf{Category} & \textbf{Feature} & \textbf{Existing Benchmarks} \\\hline
     Logical Reasoning & Fundamental logical reasoning, including inductive, deductive and abduactive reasoning tasks. & LogicInference, DEER\\\hline
     Non-text Semantic Reasoning & Non-verbal reasoning questions, such as mathematical, temporal and spatial reasoning. & GSM8k, SVAMP, AQuA, MultiArith, TimeDial, SpartQA, StepGame\\\hline
     Commonsense Reasoning & Implicit commonsense knowledge to solve the problem. & CommonsenseQA, OpenbookQA, AI2 Reasoning Challenge, BoolQ\\\hline
     Complex Reasoning & Challenging tasks or realistic examination problems. & ReClor, LogiQA\\\hline
    \end{tabular}}
    \caption{Main categories of reasoning tasks and benchmarks}
    \label{tab:reason}
    
\end{table*}

The foremost among these is logical reasoning, encompassing inductive, deductive, and abductive reasoning. It is a logic process fundamentally rooted in deriving conclusions or judgments based on given evidence or experience~\cite{bhagavatula2019abductive,sinha2019clutrr,clark2018think}.

Another notable domain under scrutiny is non-text semantic reasoning, which involves mathematical, temporal, and spatial reasoning. Those reasoning tasks are intended to test the understanding of non-verbal questions and the logic's coherence, completeness, and correctness. Prominent benchmarks in mathematical reasoning include the grade school math problem GSM8K~\citep{cobbe2021training}, math word problems SVAMP~\citep{patel2021nlp}, and two algebraic word problems AQuA~\citep{ling2017program}, MultiArith~\citep{roy2016solving}. For temporal reasoning, ~\citet{qin2021timedial} is designed to understand the time duration and the relation between events. SpartQA~\citet{mirzaee2021spartqa} and StepGame~\citet{shi2022stepgame} are two pioneering benchmarks in the field of spatial reasoning evaluating baselines' understanding of spatial relationships. 

Commonsense reasoning emerges as a pivotal area. This type of reasoning tasks require LLMs make some elementary judgments and prediction on daily concepts and knowledge, such as CommonsenseQA~\citep{talmor2018commonsenseqa}, OpenBookQA~\citep{mihaylov2018can}, StrategyQA~\citep{geva2021did}, AI2 Reasoning Challenge~\citep{clark2018think}, BoolQ~\citep{clark2019boolq}. 

Complex reasoning is the final yet crucial category of reasoning derived from cognitive science to evaluate LLMs. It includes understanding, causal, multi-hop, and analogical reasoning, which delves into the nuanced interplay of cause and effect, the ability to traverse multiple inferential steps for conclusions and parallels between seemingly disparate scenarios. Causal reasoning is indented to identify the relationship between causes and actions~\cite{du2022care,du2021excar}. Multi-hop reasoning is another cornerstone reasoning over a larger context~\cite{yang2018hotpotqa}. Analogical reasoning is to examine the process of thinking relying on analogy or counter facts to drive a conclusion~\cite{webb2023emergent}. 

Existing reasoning benchmarks, while valuable, have often presented a somewhat narrow view of cognitive processes, not fully capturing the breadth and depth inherent in genuine human reasoning. The necessity of a general benchmark focused on mental abilities emerges. 

Similar issues arise in the field of modeling. Mainstream reasoning methodologies involve fully supervised finetuning, prompting\&in-context Learning, and hybrid methods. Before fully utilizing LLMs in reasoning, previous works are also notable. For example, ~\citet{rajani2019explain} finetuned a pre-trained GPT model in order to generate rationales that explain model predictions. ~\cite{talmor2020leap} train models for multi-step reasoning for program synthesis. According to~\cite{huang2022towards}, the limits are bi-fold: Firstly, fully supervised finetuning demands a dataset that encapsulates explicit reasoning, a task that is both challenging and time-intensive to curate. Secondly, when the model is rigorously trained on this specific dataset, it inherently becomes confined to that particular domain instead of a general reasoning ability. 

Recently, prompting to simulate a human-like cognitive process has become a remarkable area, such as CoT and its variants~\cite{wei2022chain,wang2022self,zelikman2022star,yao2023tree,yao2023beyond}, rationale engineering~\cite{fu2022complexity,kojima2022large} as well as question decomposition~\cite{zhou2022least, drozdov2022compositional,khot2022decomposed}. Although recent research indicates that these models fall short when they require multiple reasoning steps, delving into LLMs' capacity for human-like logical reasoning has unmistakably emerged as a prevailing trend in the field. Hence, we believe it is imperative to introduce new benchmarks and tasks that represent a high level of human logic. 

\section{Conclusions and Future Work}
 In this work, we propose IEP, a novel prompting framework inspired by the human decision-making process - forward proposing several options and then eliminating those contradicted with the context.
 We inventively align this reasoning process with NLI tasks. 
 We also introduce a general and challenging benchmark in the field of complex reasoning, \textsc{MARB}, including several innovative tasks. 
Empirical results show that IEP can serve as a promising framework for guiding LLMs on some structured and rigorous reasoning tasks. Its' forward planning and backward inferring can help LLMs perform well in many tasks, especially where CoT struggles.

We envision future work in two directions. The first is to build more corresponding benchmarks with IEP human rationale reference, which is a stepping stone for further improving LLMs' thinking back and eliminating abilities. 
We envision future work in two directions. The first is to build more corresponding benchmarks with IEP human rationale reference, which is a stepping stone for further improving LLMs' thinking back abilities. 
The second of which is inspired by CoT's further improvements. The settings of both forward reasoning and backward inferring are somehow a type of backtracking. 
Some algorithms and data structures, such as trees or graphs, can work on that. So, this work provides a direction for future advanced work.

\section*{Limitations}

\textbf{Active retrieval}~ we limit our experiments to reason without domain knowledge in this work. However, in real-world human reasoning, when we make eliminating decisions, it is common to search for related knowledge. In future works, we aim to extend our prompting framework with retrieval-augmented skills. This potential strategy could be helpful in the correctness of both inferring and eliminating. 
\\
\textbf{Shortage in human elimination reference}~Because IEP is a new framework, previous work has limited attention to creating large-scale human elimination thinking references. This might be pretty helpful if we want to guide LLMs to refer to eliminate incorrect options.

% \section*{Ethics Statement}

\section*{Acknowledgements}
We are grateful to Google for making PaLM2 API available for free. 
% Entries for the entire Anthology, followed by custom entries
\bibliography{anthology,custom}
\bibliographystyle{acl_natbib}

\appendix
\section{Examples in \textsc{MARB}}
\label{app:examples}
\begin{figure}[htbp]
    \centering
    \scalebox{0.24}{\includegraphics{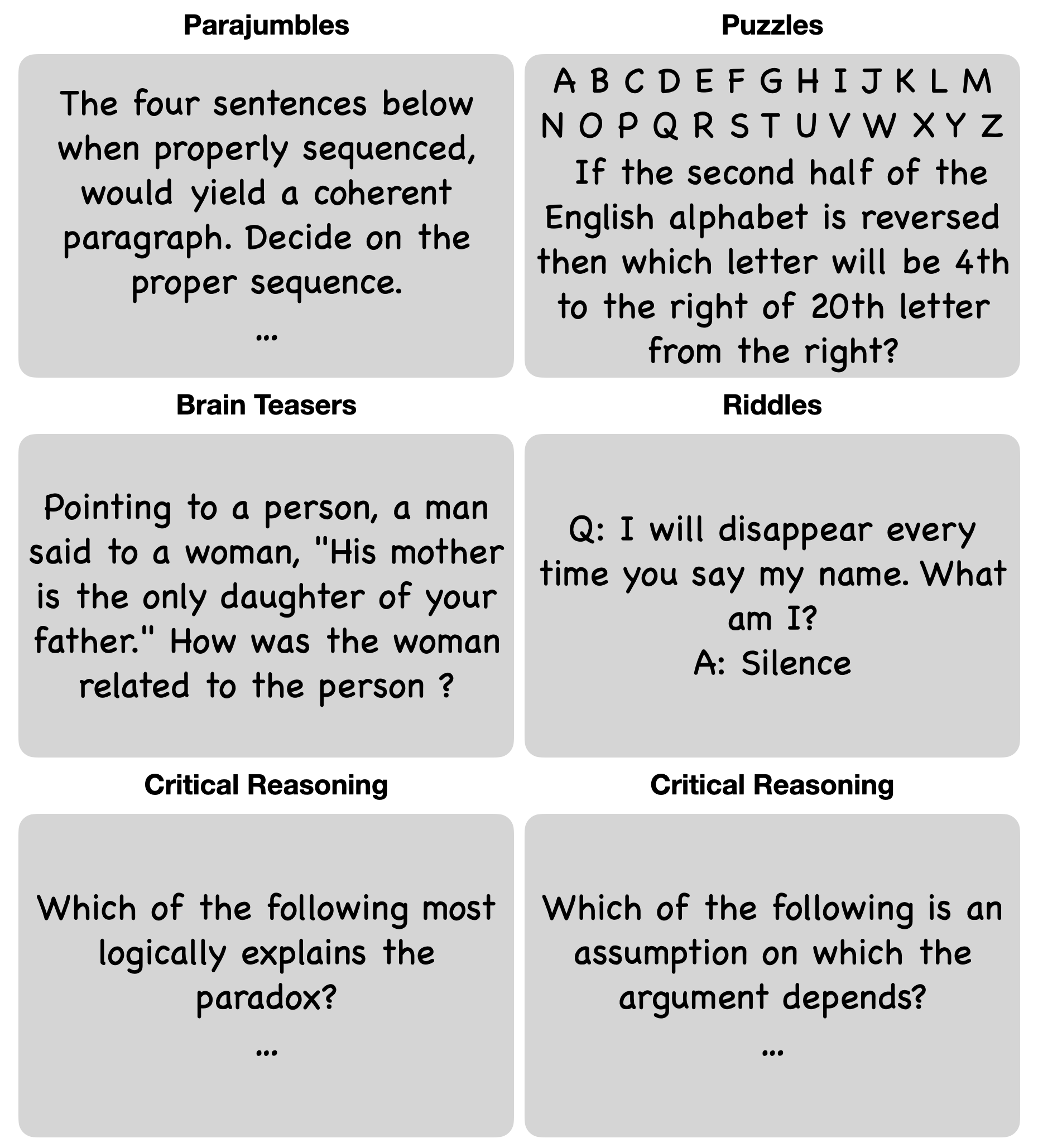}}
    \caption{Examples from \textsc{MARB}}
    \label{fig:benchmarks}
\end{figure}

\end{document}